\documentclass{article}

\usepackage[accepted]{icml2019}
\usepackage{color}
\usepackage{tikz,pgfplots}
\usepackage{amsmath,graphicx}
\usepackage{amsthm,amssymb}
\usepackage{algorithmicx}
\usepackage{algpseudocode}

\newtheorem{assumption}{Assumption}
\newenvironment{proof-sketch}{\noindent{\bf Sketch of Proof}\hspace*{1em}}{\qed\bigskip}

\usepackage{color}
\usepackage{float}
\usepackage{appendix}
\usepackage{comment}
\theoremstyle{definition}

\newcommand{\trans}{{\sf T}}
\newcommand{\EE}{{\rm E}}
\newcommand{\tr}{{\rm tr}}
\newcommand{\RR}{{\mathbb{R}}}
\newcommand{\CC}{{\mathbb{C}}}
\usepackage[colorinlistoftodos]{todonotes}
\DeclareMathOperator{\argmin}{argmin}
\DeclareMathOperator{\diag}{diag}

\newtheorem{remark}{Remark}
\newtheorem{proposition}{Proposition}

\icmltitlerunning{Random matrix improved covariance estimation}

\begin{document}

\twocolumn[
\icmltitle{Random Matrix Improved Covariance Estimation for a Large Class of Metrics}




\begin{icmlauthorlist}
\icmlauthor{Malik Tiomoko}{CS,GIPSA}
\icmlauthor{Florent Bouchard}{GIPSA}
\icmlauthor{Guillaume Ginholac}{LISTIC}
\icmlauthor{Romain Couillet}{GIPSA,CS}
\end{icmlauthorlist}

\icmlaffiliation{CS}{CentraleSup\'elec, University ParisSaclay, France}
\icmlaffiliation{GIPSA}{GIPSA-lab, University Grenoble-Alpes, France}
\icmlaffiliation{LISTIC}{LISTIC, University Savoie Mont-Blanc, France}

\icmlcorrespondingauthor{Malik Tiomoko}{malik.tiomoko@gipsa-lab.grenoble-inp.fr}
\icmlcorrespondingauthor{Florent Bouchard}{florent.bouchard@gipsa-lab.grenoble-inp.fr}
\icmlcorrespondingauthor{Guillaume Ginholac}{guillaume.ginholac@univ-smb.fr}
\icmlcorrespondingauthor{Romain Couillet}{romain.couillet@gipsa-lab.grenoble-inp.fr}
\icmlkeywords{covariance matrix, precision matrix, LDA, QDA}

\vskip 0.3in
]
\printAffiliationsAndNotice{}


\begin{abstract}
Relying on recent advances in statistical estimation of covariance distances based on random matrix theory, this article proposes an improved covariance and precision matrix estimation for a wide family of metrics. 
The method is shown to largely outperform the sample covariance matrix estimate and to compete with state-of-the-art methods, while at the same time being computationally simpler. Applications to linear and quadratic discriminant analyses also demonstrate significant gains, therefore suggesting practical interest to statistical machine learning.
\end{abstract}

\section{Introduction}
\label{sec:intro}
Covariance and precision matrix estimation is a fundamental and simply posed, yet still largely considered, key problems of statistical data analysis, with countless applications in statistical inference. In machine learning, it is notably at the core of elementary methods as linear (LDA) and quadratic discriminant analysis (QDA) \cite{mclachlan2004discriminant}.

Estimation of the covariance matrix $C\in\RR^{p\times p}$ based on $n$ independent (say zero mean) samples $x_1,\ldots,x_n\in\RR^p$ is conventionally performed using the sample covariance matrix (SCM) $\hat C\equiv \frac1n\sum_{i=1}^nx_ix_i^\trans$ (and its inverse using $\hat C^{-1}$). The estimate is however only consistent for $n\gg p$ and only invertible for $n\geq p$. Treating the important practical cases where $n\sim p$ and even $n\ll p$ has recently spurred a series of parallel lines of research. These directions rely either on structural constraints, such as ``toeplitzification'' procedures for Toeplitz covariance models (particularly convenient for time series) \cite{BIC08,WU09,VIN14}, on sparse constraints with LASSO and graphical LASSO-based approaches \cite{FRI08} or, more interestingly for the present article, on exploiting the statistical independence in the entries of the vectors $x_i$.

\cite{LED04} proposes to linearly ``shrink'' $\hat C$ as $\hat C(\rho)\equiv\rho I_p+\sqrt{1-\rho^2}\hat C$ for $\rho>0$ chosen to minimize the expected Frobenius distance $\EE[\|C-\hat C(\rho)\|_F]$ in the asymptotic $p,n\to\infty$ limit with $p/n\to c>0$. Basic results from random matrix theory (RMT) are used here to estimate $\rho$ consistently. This procedure is simple and quite flexible and has been generalized in various directions (e.g., in \cite{couillet2014large} with a robust statistics approach). However, the method only applies a naive homothetic map to each $\lambda_i(\hat C)$ of $\hat C$ in order to better estimate $\lambda_i(C)$. A strong hope to recover a better approximation of the $\lambda_i(C)$'s then arose from \cite{SIL95,CHO95} that provide a random matrix result relating directly the limiting eigenvalue distributions of $C$ and $\hat C$. Unfortunately, while estimating the $\lambda_i(\hat C)$'s from the $\lambda_i(C)$'s is somewhat immediate, estimating the $\lambda_i(C)$'s backward from the $\lambda_i(\hat C)$'s is a difficult task. \cite{ELK08} first proposed an optimization algorithm to numerically solve this problem, however with little success as the method is quite unstable and has rarely been efficiently reproduced. \cite{MES08} later offered a powerful idea, based on contour integral, to consistently estimate linear functionals $\frac1n\sum_{i=1}^nf(\lambda_i(C))$ from the $\lambda_i(\hat C)$'s. But $f$ is constrained to be very smooth (complex analytic) which prevents the estimation of the individual $\lambda_i(C)$'s. Recently, Ledoit and Wolf took over the work of El~Karoui, which they engineered to obtain a more efficient numerical method, named QuEST \cite{LW15}. Rather than inverting the Bai--Silverstein equations, the authors also proposed, with the same approach, to estimate the $\lambda_i(C)$'s by minimizing a Frobenius norm distance \cite{LW15} (named QuEST1 in the present article) or a Stein loss \cite{LW18} (QuEST2 here).

These methods, although more stable than El~Karoui's initial approach, however suffer several shortcomings: (i) they are still algorithmically involved as they rely on a series of fine-tuned optimization schemes, and (ii) they are only adaptable to few error metrics (Frobenius, Stein).

Inspired by Mestre's approach and the recent work \cite{couillet2018random}, this article proposes a different procedure consisting in (i) writing $C$ as the solution to $\argmin_{M\succ 0}\delta(M,C)$ for a wide range of metrics $\delta$ (Fisher, Batthacharyya, Stein's loss, Wasserstein, etc.), (ii) based on \cite{couillet2018random}, using the fact that $\delta(M,C)-\hat\delta(M,X)\to 0$ for some consistent estimator $\hat\delta$, valid for all deterministic $M$ and samples $X=[x_1,\ldots,x_n]\in\RR^{p\times n}$ having zero mean and covariance $C$, and (iii) proceeding to a gradient descent on $\hat{\delta}$ rather than on the unknown $\delta$ itself. With appropriate adaptations, the estimation of $C^{-1}$ is similarly proposed by solving instead $\argmin_{M\succ 0}\delta(M,C^{-1})$.

While only theoretically valid for matrices $M$ independent of $X$, and thus only in the first steps of the gradient descent, the proposed method has several advantages: (i) it is easy to implement and technically simpler than QuEST, (ii) it is adaptable to a large family of distances and divergences, and, most importantly, (iii) simulations suggest that it systematically outperforms the SCM and is competitive with, if not better than, QuEST.

\medskip

The remainder of the article is organized as follows. Section~\ref{sec:model} introduces preliminary notions and concepts on which are hinged our proposed algorithms, thereafter described in Section~\ref{sec:methodology}. Section~\ref{sec:simu} provides experimental validations and applications, including an improved version of LDA/QDA based on the proposed enhanced estimates.

\medskip

\noindent{\bf Reproducibility.} Matlab codes for the proposed estimation algorithms are available as supplementary materials and are based on Manopt, a Matlab toolbox for optimization on manifolds \cite{manopt}.

\section{Preliminaries}
\label{sec:model}
Let $n$ random vectors $x_1,\ldots,x_n\in\RR^p$ be of the form ${x_i=C^{\frac12}z_i}$ for some positive definite matrix ${C\in\RR^{p\times p}}$ and $z_1,\ldots,z_n\in\RR^p$ independent random vectors of independent entries with $\EE[ [z_i]_j]=0$ and $\EE[ |[z_i]_j|^2]=1$. We further assume the following large dimensional regime for $n$ and $p$.

\begin{assumption}[Growth Rate]
\label{ass:growth rates}
As $n \to\infty$, ${p/n\to c\in (0,1)}$ and $\limsup_p\max\{\|C^{-1}\|,\|C\|\} < \infty$ for $\|\cdot\|$ the matrix operator norm.
\end{assumption}

Our objective is to estimate $C$ and $C^{-1}$ based on $x_1,\ldots,x_n$ under the above large $p,n$ regime. For simplicity of exposition and readability, we mostly focus on the estimation of $C$ and more briefly discuss that of $C^{-1}$. 

Our approach relies on the following elementary idea: 
\begin{equation*}
    C \equiv \argmin_{M \succ 0} \delta(M,C)
\end{equation*}
where, for some function $f$,
\begin{align}
\label{eq:delta}
    \delta(M,C) &\equiv \frac1p\sum_{i=1}^p f(\lambda_i(M^{-1}C))
\end{align}
is a divergence (possibly a squared distance $\delta=d^2$) between the positive definite matrices $M$ and $C$, depending only on the eigenvalues of $M^{-1}C$.
Among divergences satisfying this condition, we find the \emph{natural Riemannian distance} $d_{\textup{R}}^2$~\cite{B09}, which corresponds to the Fisher metric for the multivariate normal distribution~\cite{S84}; the \emph{Battacharyya distance} $d_{\textup{B}}^2$~\cite{S13}, which is close to the natural Riemannian distance while numerically less expensive; the \emph{Kullback-Leibler divergence} $\delta_{\textup{KL}}$, linked to the likelihood and studied for example in~\cite{M12}; the \emph{R\'enyi divergence} $\delta_{\alpha\textup{R}}$ for Gaussian $x_i$'s~\cite{van2014renyi}; etc.\footnote{The Frobenius distance does not fall into this setting but has already largely been investigated and optimized.} Table~\ref{tab:f} reports the explicit values of $f$ for these divergences. 

Since $\delta(M,C)$ is not accessible as $C$ is unknown, our approach exploits an estimator for $\delta(M,C)$ which is consistent in the large $n,p$ regime of Assumption~\ref{ass:growth rates}.

\medskip

To this end, our technical arguments are fundamentally based on random matrix theory, and notably rely on the so-called \emph{Stieljes transform} of eigenvalue distributions. For an arbitrary real-supported probability measure $\theta$, the Stieltjes transform $m_{\theta}:\CC\setminus {\rm supp}(\theta)\to \CC$ is defined as
\begin{align*}
    m_{\theta}(z)= \int\frac{\theta(dt)}{t-z}.
\end{align*}
The key interest of the Stieltjes transform in this article is that it allows one to relate the distributions of the eigenvalues of $C$ and $\hat C$ as $p,n\to\infty$ \cite{SIL95}. More specifically here, for arbitrary deterministic matrices $M$, Stieltjes transform relations connect the empirical spectral (i.e., eigenvalue) distribution $\nu_p$ of $M^{-1}C$ to the empirical spectral distribution $\mu_p$ of $ M^{-1}\hat C$ \cite{couillet2018random}, defined as
\begin{align*}
    \mu_p\equiv\frac1p\sum_{i=1}^p \delta_{\lambda_i(M^{-1}\hat{C})}~\textmd{ and }~ \nu_p\equiv\frac1p\sum_{i=1}^p \delta_{\lambda_i(M^{-1}C)}.
\end{align*}

\medskip

The connecting argument goes as follows: first, from Cauchy's integral formula (stating that $f(t)=\frac1{2\pi\imath}\oint_\Gamma f(z)/(t-z)dz$ for $\Gamma$ a complex contour enclosing $t$), the metric $\delta(M,C)$ in \eqref{eq:delta} relates to the Stieltjes transform $m_{\nu_p}(z;M)$ through
\begin{equation}
\label{eq:Cauchy}
    \delta(M,C)=\frac1{2\pi\imath}\oint_{\Gamma} f(z)m_{\nu_p}(z;M)dz
\end{equation}
for $\Gamma\subset \CC$ a (positively oriented) contour surrounding the eigenvalues of $M^{-1}C$. The notation $m_{\nu_p}(z;M)$ reminds the dependence of $m_{\nu_p}(z)$ in the matrix $M$.

\begin{table}
\centering
\begin{tabular}{c|l}
Divergences & $f(z)$ \\
\hline
$d_{\textup{R}}^2$ & $\log^{2}(z)$ \\
$d_{\textup{B}}^2$ & $-\frac14\log(z)+\frac12\log(1+z)-\frac12\log(2)$ \\
$\delta_{\textup{KL}}$ & $\frac12z-\frac12\log(z)-\frac12$ \\
$\delta_{\alpha\textup{R}}$ & $\frac{-1}{2(\alpha-1)}\log(\alpha+(1-\alpha)z)+\frac12\log(z)$
\end{tabular}
\caption{Distances $d$ and divergences $\delta$, and their corresponding $f(z)$ functions.}
\label{tab:f}
\end{table}

Then, by exploiting the relation between the Stieltjes transforms $\mu_{\nu_p}$ and $m_{\mu_p}$, it is shown in \cite{couillet2018random} that, under Assumption~\ref{ass:growth rates}, for all deterministic $M$ of bounded operator norm,
\begin{equation}
\label{eq:hatdelta}
    \delta(M,C) - \hat \delta(M,X) \to 0
\end{equation}
almost surely, where $X=[x_1,\ldots,x_n]$ and
\begin{equation}
    \hat \delta(M,X) \equiv \frac{1}{2\pi \imath c} \oint_{\hat \Gamma} G\left(-m_{\tilde{\mu}_p}(z;M)\right)dz
    \label{eq:distance_integral}
\end{equation}
with $G$ such that $G'(z)\equiv g(z)=f(1/z)$, $\hat \Gamma$ a contour surrounding the support of the almost sure limiting eigenvalue distribution of $M^{-1}\hat{C}$ and $\tilde{\mu}_{p}=\frac{p}n\mu_p+(1-\frac{p}n)\delta_0$ (and thus $m_{\tilde{\mu}_{p}}(z)=cm_{\mu_p}(z)+(1-\frac{p}n)/z$). Note that, by the linearity of $G$ in \eqref{eq:distance_integral}, it is sufficient in practice to evaluate $\hat\delta(M,X)$ for elementary functions (such as $f(z)=z$, $f(z)=\log(z)$, etc.) in order to cover most distances and metrics of interest (see again Table~\ref{tab:f}). Table~\ref{tab:FG} reports the values of $G$ for such atomic functions $f$.

Our main idea is to estimate $C$ by minimizing the approximation $\hat{\delta}(M,X)$ of $\delta(M,C)$ over $M$. However, it is important to note that, as discussed in \cite{couillet2018random}, the random quantity $\hat\delta(M,X)$ may be negative with non-zero probability. As such, minimizing $\hat\delta(M,X)$ over $M$ may lead to negative solutions. Our proposed estimation method therefore consists in approximating $C$ by the solution to the optimization problem
\begin{align}
\label{eq:minimization}
    \argmin_{M\succ 0} \{h_X(M)\}, \textmd{~where~} h_X(M)\equiv(\hat\delta(M,X))^2.
\end{align}

\section{Methodology and Main Results}
\label{sec:methodology}

\subsection{Estimation Method}

We solve \eqref{eq:minimization} via a gradient descent algorithm in the Riemannian manifold $S_n^{++}$ of positive definite $n\times n$ matrices.

To evaluate the gradient $\nabla h_X(M)$ of $h_X$ at $M\in S_n^{++}$, recall that on $S_n^{++}$ the differential ${\rm D}h_X(M)[\xi]$ of the functional $h_X:S_n^{++}\to \RR^{+}$, at position $M\in S_n^{++}$ and in the direction of $\xi\in S_n$ (the Riemannian manifold of symmetric $n\times n$ matrices), is given by \cite{absil2009optimization}
\begin{equation*}
    {\rm D}h_X(M)[\xi]=\langle \nabla h_X(M),\xi\rangle_{M}^{S_n^{++}}
\end{equation*}
where $\langle\cdot,\cdot\rangle_{.}^{S_n^{++}}$ is the Riemannian metric defined through
\begin{equation*}
    \langle\eta,\xi\rangle_{M}^{S_{n}^{++}}=\tr\left(M^{-1}\eta M^{-1} \xi \right).
\end{equation*}

Differentiating $\hat\delta^2(M,X)$ at $M$ in the direction $\xi$ yields:
\begin{align*}
    &{\rm D} h_X(M)[\xi]\\
    &=\frac{-\hat\delta(M,X)}{\pi i c}\oint_{\Gamma}g(-m_{\tilde{\mu_p}}\left(z,M\right)){\rm D}m_{\tilde{\mu_p}}\left(z,M\right)[\xi]dz.
\end{align*}
By using the fact that
\begin{align*}
    &{\rm D}m_{\tilde{\mu_p}}\left(z,M\right)[\xi] \\
    &=\frac{c}{p}{\rm D} \tr\left(\left[M^{-1}\hat{C}-zI_{p}\right]^{-1}\right)[\xi]\\
    &=\frac{c}{p}\tr\left(M^{-1}\hat{C}\left[M^{-1}\hat{C}-zI_{p}\right]^{-2}M^{-1}\xi\right)\\
    &=\left\langle \frac{c}{p}{\rm sym} \left(\hat{C}\left[M^{-1}\hat{C}-zI_{p}\right]^{-2}\right),\xi\right\rangle_{M}^{S_n^{++}}
\end{align*}
where ${\rm sym}(A)=\frac12(A+A^\trans)$ is the symmetric part of $A\in\RR^{p\times p}$, we retrieve the gradient of $h_X(M)$ as
\begin{align}
    &-\imath\pi p\frac{\nabla h_X(M)}{\hat\delta(M,X)} \nonumber \\ \label{eq:integral_gradient}
    &=\oint_{\hat\Gamma} g\left(-m_{\tilde{\mu}_p}(z;M)\right){\rm sym}\left(\hat{C}(M^{-1}\hat{C}-zI_{p})^{-2}\right)dz
\end{align}
(recall that the right-hand side still depends on $X$ implicitly through $\tilde{\mu}_p$ and $\hat{C}$).

\begin{table}[t]
\centering
\begin{tabular}{l|l}
$f(z)$ & $G(z)$ \\
\hline
$\log^{2}(z)$ & $z\left(\log^{2}(z)-2\log(z)+2\right)$ \\
$\log(z)$ & $-z\log(z)+z$ \\
$\log(1+sz)$ & $s\log(s+z)+z\log\left(\frac{s+z}{z}\right)$\\
$z$ & $\log(z)$
\end{tabular}

\medskip

\begin{tabular}{l|l}
$f(z)$ & $F(z)$ \\
\hline
$\log^{2}(z)$ & $z\left(\log^{2}(z)-2\log(z)+2\right)$ \\
$\log(z)$ & $z\log(z)-z$ \\
$\log(1+sz)$ & $\left(\frac1s+z\right)\log(1+sz)-z$\\
$z$ & $\frac12z^2$
\end{tabular}
\caption{Values of $G(z)$ and $F(z)$ for ``atomic'' $f(z)$ functions used in most distances and divergences under study; here $s>0$ and $z\in\CC$.}
\label{tab:FG}
\end{table}

\medskip

Once $\nabla h_X$ estimated, every gradient descent step in $S_n^{++}$ corresponds to a small displacement on the geodesic starting at $M$ and towards $-\nabla h_X(M)$, defined as the curve
\begin{align*}
    \RR_+ &\to S_n^{++} \\
    t   &\mapsto M^{\frac12} \exp\left(-t M^{-\frac12} \nabla h_X(M) M^{-\frac12} \right) M^{\frac12}
\end{align*}
where, for $A=U\Lambda U^\trans\in S_n^{++}$ in its spectral decomposition, $\exp(A)\equiv U\exp(\Lambda)U^\trans$ (with $\exp$ understood here applied entry-wise on the diagonal elements of $\Lambda$).

That is, letting $M_0,M_1,\ldots$ and $t_0,t_1,\ldots$ be the successive iterates and step sizes of the gradient descent, we have, for some given initialization $M_0\in S_n^{++}$,
\begin{align}
\label{eq:Mk+1}
    M_{k+1} &= M_k^{\frac12} \exp\left(-t_k M_k^{-\frac12} \nabla h_X(M_k) M_k^{-\frac12} \right) M_k^{\frac12}.
\end{align}

Our proposed method is summarized as Algorithm~\ref{alg:estimC}.

\begin{algorithm}
{\bf Require} $M_0\in C_n^{++}$.

\smallskip

{\bf Repeat} {$M \gets M^{\frac12} \exp\left(-t M^{-\frac12} \nabla h_X(M) M^{-\frac12} \right) M^{\frac12}$ with $t$ either fixed or optimized by backtracking line search.}

{\bf Until} {Convergence.}

\smallskip

{{\bf Return} $M$.}

\smallskip

\caption{Proposed estimation algorithm.}
\label{alg:estimC}
\end{algorithm}

We conclude this section by an important remark on the fundamental limitations of the proposed algorithm.

\begin{remark}[Approximation of $\delta(M_k,C)$ by $\hat\delta(M_k,X)$]
\label{rem:approx}
    It is fundamental to understand the result from \cite{couillet2018random} at the heart of the proposed method. There, it is precisely shown that, for every deterministic sequence of matrices $\{M^{(p)},~p=1,2,\ldots\}$ and $\{C^{(p)},~p=1,2,\ldots\}$, with $M^{(p)},C^{(p)}\in\RR^{p\times p}$ and $\max(\|C^{(p)}\|,\|M^{(p)}\|)<K$ for some constant $K$ independent of $p$, we have that, for $X^{(p)}=[x^{(p)}_1,\ldots,x^{(p)}_n]$ with $x_i^{(p)}=C^{(p)\frac12}z_i^{(p)}$ and $z_i^{(p)}$ i.i.d.\@ vectors of i.i.d.\@ zero mean and unit variance entries, 
    \begin{align*}
        \delta(M^{(p)},C^{(p)}) - \hat\delta(M^{(p)},X^{(p)}) \to 0
    \end{align*}
    almost surely as $n,p\to\infty$ and $p/n\to c\in (0,1)$. This result seems to suggest that $\hat\delta(M_k,X)$ in our algorithm is a good approximation for the sought for $\delta(M_k,C)$. This however only holds true so long that \emph{$M_k$ is independent of $X$} which clearly does not stand when proceeding to successive gradient descent steps in the direction of $\nabla h_X(M)$ which depends explicitly on $X$. As such, while initializations with, say, $M_0=I_p$, allow for a close approximation of $\delta(M_k,C)$ in the very first steps of the descent, for larger values of $k$, the descent is likely to drive the optimization in less accurate directions. 
    
\end{remark}

\begin{figure}
\begin{tabular}{cc}
\hspace{-0.25cm}
    \begin{tikzpicture}[scale=.51]
    		\begin{axis}[grid=major,xlabel={algorithm step $k$},xmin=1,xmax=19,ymin=0]
    			            \addplot[thin,mark=x,mark size=3pt,blue,thick]coordinates{            (1,3.774285e-01)(2,2.337737e-01)(3,9.137719e-02)(4,1.317583e-02)(5,3.126441e-03)(6,8.535357e-04)(7,5.785198e-06)(8,7.132698e-08)(9,9.165165e-10)(10,1.183026e-11)(11,1.527808e-13)(12,1.973187e-15)(13,2.548089e-17)(14,3.306143e-19)(15,4.181741e-21)(16,5.778073e-23)(17,2.711229e-27)(18,2.397163e-27)(19,2.397163e-27)

};
    			            \addplot[thin,mark=o,mark size=3pt,red,thick]coordinates{            (1,3.791677e-01)(2,3.007303e-01)(3,2.144808e-01)(4,1.502759e-01)(5,1.349888e-01)(6,1.287918e-01)(7,1.230254e-01)(8,1.225852e-01)(9,1.225366e-01)(10,1.225311e-01)(11,1.225305e-01)(12,1.225304e-01)(13,1.225304e-01)(14,1.225304e-01)(15,1.225304e-01)(16,1.225304e-01)(17,1.225304e-01)(18,1.225304e-01)(19,1.225304e-01)

};
    			            \legend{{$\hat\delta(M_k,X)$},{$\delta(M_k,C)$}};
    		\end{axis}
    \end{tikzpicture}& 
    \hspace{-0.5cm}
    \begin{tikzpicture}[scale=.51]
            \begin{axis}[grid=major,scaled ticks=false,xlabel={algorithm step $k$},xmin=1,xmax=19,ymin=0, tick label style={/pgf/number format/fixed}]
    			            \addplot[ultra thin,mark=x,mark size=1pt,blue,thick]coordinates{        (1,1.739173e-03)(2,6.695659e-02)(3,1.231037e-01)(4,1.371001e-01)(5,1.318624e-01)(6,1.279382e-01)(7,1.230196e-01)(8,1.225851e-01)(9,1.225366e-01)(10,1.225311e-01)(11,1.225305e-01)(12,1.225304e-01)(13,1.225304e-01)(14,1.225304e-01)(15,1.225304e-01)(16,1.225304e-01)(17,1.225304e-01)(18,1.225304e-01)(19,1.225304e-01)

};
    			           \legend{{$\delta(M_k,C)-\hat\delta(M_k,X)$},};
    		\end{axis}
    \end{tikzpicture}
\end{tabular}
\caption{(left) Evolution of the Fisher distance $\delta(M_k,C)$ versus $\hat\delta(M_k,X)$ for $k=1,2,\ldots$, initialized to $M_0=I_p$,. (right) Evolution of $\delta(M_k,C)-\hat\delta(M_k,X)$.}
\label{fig:cost_distance}
\end{figure}
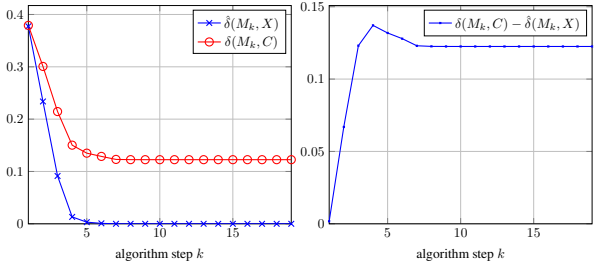

Remark~\ref{rem:approx} is in fact not surprising. Indeed, finding the minimum of $\delta(M,C)$ over $M\succ 0$ would result in finding $C$, which cannot be achieved for unconstrained matrices $C$ and for non vanishing values of $p/n$. Figure~\ref{fig:cost_distance} provides a typical evolution of the distance $\delta(M_k,C)$ versus its approximation $\hat\delta(M_k,X)$ at the successive steps $k=1,2,\ldots$ of Algorithm~\ref{alg:estimC}, initialized at $M_0=I_p$. As expected, the difference $|\hat\delta(M_k,X)-\delta(M_k,C)|$, initially small (at $k=1$, $\hat\delta(I_p,X)\simeq \delta(I_p,C)$), increases with $k$, until the gradient vanishes and the divergence $\delta(M_k,C)$ converges.

\subsection{Practical Implementation}
\label{sec:practical}
In order to best capture the essence of Algorithm~\ref{alg:estimC}, as well as its various directions of simplification and practical fast implementation, a set of important remarks are in order.

\medskip

    First note that the generic computation of $M_{k+1}$ in Equation~\eqref{eq:Mk+1} may be numerically expensive, unless $M_k$ and $\nabla h_X(M)$ share the same eigenvectors. In this case, letting $M_k=U\Omega_k U^\trans$ and $\nabla_Xh(M_k)=U\Delta_k U^\trans$, we have the recursion
    \begin{equation}
    \label{eq:recursion_Mk}
        [\omega_{k+1}]_i = [\omega_k]_i \exp\left( -t \frac{[\delta_k]_i}{[\omega_k]_i} \right)
    \end{equation}
    where $\delta_k=\diag(\Delta_k)$ and $\omega_k=\diag(\Omega_k)$.
    
    In particular, if $M_0=\alpha I_n+\sqrt{1-\alpha^2}\hat C$ is a linear shrinkage for some $\alpha\in [0,1]$, we immediately find that, for all $k\geq 0$,
    \begin{align*}
        M_k &= \hat U \Omega_k \hat U^\trans
    \end{align*}
    where $\hat U\in\RR^{p\times p}$ are the eigenvectors of $\hat C$ (i.e., in its spectral decomposition, $\hat C=\hat U \hat\Lambda\hat U^\trans$) and $\Omega_k$ is recursively defined through \eqref{eq:recursion_Mk}.
    
    This shows that, initialized as such, the ultimate limiting estimate $M_\infty$ (i.e., the limit of $M_k$) of $C$ shares the same eigenvectors as $\hat C$, and thus reduces to a ``non-linear shrinkage'' procedure, similar to that of \cite{LW15}. Extensive simulations in fact suggest that, if initialized randomly (say with $M_0$ a random Wishart matrix), after a few iterates, the eigenvectors of $M_k$ do converge to those of $\hat C$ (see further discussions in Section~\ref{sec:simu}). As such, for computational ease, we suggest to initialize the algorithm with $M_0=I_p$ or with $M_0$ a linear shrinkage of $\hat C$.

\medskip

A further direction of simplification of Algorithm~\ref{alg:estimC} relates to the fact that, for generic values of $M_0$ (notably having eigenvectors different from those of $\hat C$), Equation~\eqref{eq:Mk+1} is computationally expensive to evaluate. A second-order simplification for small $t$ is often used in practice \cite{JVV12}, as follows
\begin{align*}
    M_{k+1} &= M_k - t \nabla h_X(M_k) \nonumber \\
    &+ \frac{t^2}2 \nabla h_X(M_k) M_k^{-1}\nabla h_X(M_k) +O(t^3).
\end{align*}
Simulations with this approximation suggest almost no difference in either the number of steps until convergence or accuracy of the solution.

\subsection{Estimation of $C^{-1}$}
\label{rem:inv}
In our framework, estimating $C^{-1}$ rather than $C$ can be performed by minimizing $\delta(M,C^{-1})$ instead of $\delta(M,C)$. In this case, under Assumption~\ref{ass:growth rates}, \eqref{eq:hatdelta} now becomes
\begin{equation*}
    \delta(M,C^{-1}) - \hat \delta^{\rm inv}(M,X) \to 0
\end{equation*}
almost surely, for every deterministic $M$ of bounded operator norm and $X=[x_1,\ldots,x_n]$, where
\begin{equation*}
    \hat \delta^{\rm inv}(M,X) \equiv \frac{1}{2\pi \imath c} \oint_{\hat \Gamma} F\left(-m_{\tilde{\mu}_p^{\rm inv}}(z;M)\right)dz
\end{equation*}
for $F$ such that $F'(z)\equiv f(z)$, $\hat \Gamma$ a contour surrounding the support of the almost sure limiting eigenvalue distribution of $M\hat{C}$ and $\tilde{\mu}_{p}^{\rm inv}=\frac{p}{n}\mu_p^{\rm inv}+(1-\frac{p}{n})\delta_0$, where $\mu_p^{\rm inv}\equiv\frac1p\sum_{i=1}^p \delta_{\lambda_i(M\hat{C})}$. The cost function to minimize under this setting is now given by $h^{\rm inv}(M)\equiv(\hat\delta^{\rm inv}(M,X))^{2}$ with gradient $\nabla h_X^{\rm inv}(M)$ satisfying
\begin{align*}
    & \imath \pi p\frac{\nabla h_X^{\rm inv}(M)}{\hat\delta^{\rm inv}(M,X)} \nonumber \\ 
    &=\oint_{\hat\Gamma} f\left(-m_{\tilde{\mu}_p^{\rm inv}}(z;M)\right){\rm sym}\left(M\hat{C}(M\hat{C}-zI_{p})^{-2}M\right)dz.
\end{align*}
With these amendments, Algorithm~\ref{alg:estimC} can be adapted to the estimation of $C^{-1}$. Table~\ref{tab:FG} provides the values of $F$ for the atomic functions $f$ of interest.

\subsection{Application to Explicit Metrics}

Algorithm~\ref{alg:estimC} is very versatile as it merely consists in a gradient descent method for various metrics $f$ through adaptable definitions of the function $h_X(M)=\hat \delta(M,X)^2$ and its resulting gradient. Yet, because of the integral form assumed by the gradient (Equation~\eqref{eq:integral_gradient}), a possibly computationally involved complex integration needs to be numerically performed at each gradient descent step.

In this section, we specify closed-form expressions for the gradient for the atomic $f$ functions of Table~\ref{tab:FG} (which is enough to cover the list of divergences in Table~\ref{tab:f}). 

\subsubsection{Estimation of $C$}

Let us denote
\begin{equation*}
    \nabla h_X(M) \equiv 2\hat\delta(M,X) \cdot {\rm sym} \left( \hat C \cdot V\Lambda_\nabla V^{-1} \right)
\end{equation*}
where $V$ are the eigenvectors of $M^{-1}\hat C$ and we will determine $\Lambda_\nabla$ for each function $f$. Again, we recall from the discussion in Section~\ref{sec:practical}, that $V=\hat U$ the eigenvectors of $\hat C$ if $M$ shares the same eigenvectors as $\hat C$ (which thus avoids evaluating the eigenvectors $V$ of $M_k^{-1}\hat C$ at each step $k$ of the algorithm). 

For readability in the following, let us denote $\lambda_i\equiv \lambda_i(M^{-1}\hat C)$, $i\in\{1,\ldots,p\}$, the eigenvalues of the matrix $M^{-1}\hat{C}$ and $\xi_1,\ldots,\xi_p$ the eigenvalues of
\begin{equation*}
    \Lambda-\frac{1}{n}\sqrt{\lambda}\sqrt{\lambda}^\trans
\end{equation*}
with $\Lambda={\rm diag}(\lambda_1,\ldots,\lambda_p)$ and $\lambda=(\lambda_1,\ldots,\lambda_p)^\trans$.
Finally, for $s>0$, let $\kappa_s\in (-1/(s(1-p/n)),0)$ be the unique negative number $t$ solution of the equation (see \cite{couillet2018random} for details) 
\begin{align*}
    m_{\tilde{\mu}_p}(t)= -s.
\end{align*}

With these notations at hand, following the derivations in \cite{couillet2018random} (detailed in supplementary material), we have the following determinations for $\Lambda_\nabla$.
\begin{proposition}[Case $f(t)=t$]
For $f(t)=t$, 
\begin{align*}
    [\Lambda_\nabla]_{kk} &=-\frac1c+\frac1p \sum_{i=1}^{p}\frac{1}{m_{\tilde{\mu_p}}'(\xi_i)\left(\lambda_k-\xi_i\right)^{2}}
\end{align*}
with $m_{\tilde{\mu_p}}'$ the derivative of $m_{\tilde{\mu_p}}$.
\end{proposition}

\begin{proposition}[Case $f(t)=\log(t)$]
For $f(t)=\log(t)$,
\begin{align*}
    [\Lambda_\nabla]_{kk} &=\frac{-1}{p\lambda_k}.
\end{align*}
\end{proposition}

\begin{proposition}[Case $f(t)=\log(1+st)$]
For $s>0$ and $f(t)=\log(1+st)$,
\begin{align*}
    [\Lambda_\nabla]_{kk} &=\frac{-1}{p(\lambda_k-\kappa_s)}.
\end{align*}
\end{proposition}

\begin{proposition}[Case $f(t)=\log^{2}(t)$]
For $f(t)=\log^2(t)$,
\begin{align*}
    [\Lambda_\nabla]_{kk} &=\frac{2}{p}\log\left(\lambda_k\right)\left[\sum_{i=1}^p\frac{1}{\lambda_k-\xi_i}-\sum_{\substack{i=1\\i\neq k}}^p\frac{1}{\lambda_k-\lambda_i}-\frac{1}{\lambda_k}\right]\\
    &-\frac{2}{p}\sum_{i=1}^p\frac{\log(\xi_i)}{\lambda_k-\xi_i}+\frac{2}{p}\sum_{\substack{i=1\\i\neq k}}^p\frac{\log(\lambda_i)}{\lambda_k-\lambda_i}-\frac{2-2\log(1-c)}{p\lambda_k}.
\end{align*}
\end{proposition}

These results are mostly achieved by residue calculus for entire analytic functions $f$ or by exploiting more advanced complex integration methods (in particular branch-cut methods) for more challenging functions (involving logarithms in particular).

Combining these formulas provides an analytical expression for the gradient of all aforementioned divergences and square distances, for the estimation of $C$.

\subsubsection{Estimation of $C^{-1}$}
Similarly, for the problem of estimating $C^{-1}$, recalling Remark~\ref{rem:inv}, we may denote
\begin{equation*}
    \nabla h^{\rm inv}_X(M) \equiv 2\hat\delta^{\rm inv}(M,X) \cdot {\rm sym} \left( M \cdot V_{\rm inv}\Lambda^{\rm inv}_\nabla V_{\rm inv}^{-1} \right)
\end{equation*}
with $V_{\rm inv}$ the eigenvectors of $M\hat C$.

We redefine in this section $\lambda_i\equiv \lambda_i(M\hat C)$, and $\xi_1,\ldots,\xi_p$ the eigenvalues of $\Lambda-\frac{1}{n}\sqrt{\lambda}\sqrt{\lambda}^\trans$
with $\Lambda={\rm diag}(\lambda_1,\ldots,\lambda_p)$ and $\lambda=(\lambda_1,\ldots,\lambda_p)^\trans$.
Again, for $s>0$, let $\kappa_s<0$ be the only negative real number $t$ solution of
\begin{align*}
    m_{\tilde{\mu}_p^{\rm inv}}(t)= -\frac 1s.
\end{align*}
With the same approach as in the previous section, we here obtain the following values for $\Lambda^{\rm inv}_\nabla$.
\begin{proposition}[Case $f(t)=t$] For $f(t)=t$,
\begin{equation*}
[\Lambda_\nabla^{\rm inv}]_{kk}=-\frac{1-c}{p\lambda_k}.
\end{equation*}
\end{proposition}
\begin{proposition}[Case $f(t)=\log(t)$] For $f(t)=\log(t)$,
\begin{equation*}
[\Lambda_\nabla^{\rm inv}]_{kk}= -1 
\end{equation*}
\end{proposition}
\begin{proposition}[Case $f(t)=\log(1+st)$] For $s>0$ and $f(t)=\log(1+st)$,
\begin{equation*}
[\Lambda_\nabla^{\rm inv}]_{kk}=\frac{\lambda_k}{\lambda_k-\kappa_{s}}- 1 
\end{equation*}
\end{proposition}
\begin{proposition}[Case $f(t)=\log^{2}(t)$] For $f(t)=\log^2(t)$,
\begin{align*}
&[\Lambda_\nabla^{\rm inv}]_{kk}
=-\frac{2}{p}\log\left(\lambda_k\right)\left[\sum_{\substack{i=1\\i\neq k}}^p\frac{\lambda_k}{\lambda_k-\xi_i}-\sum_{\substack{i=1\\i\neq k}}^p\frac{\lambda_k}{\lambda_k-\lambda_i}-1\right]\\
    &+\frac{2}{p}\sum_{i=1}^p\frac{\lambda_k\log(\xi_i)}{\lambda_k-\xi_i}-\frac{2}{p}\sum_{\substack{i=1\\i\neq k}}^p\frac{\lambda_k\log(\lambda_i)}{\lambda_k-\lambda_i}+\frac{2}{p}-\frac{2}{p}\log(1-c).
\end{align*}
\end{proposition}

\section{Experimental Results}
\label{sec:simu}

This section introduces experimental results on the direct application of our proposed method to the estimation of $C$ and $C^{-1}$ as well as on its use as a plug-in estimator in more advanced procedures, here in the scope of linear and quadratic discriminant analyses (LDA/QDA).


\subsection{Validation on synthetic data}
\label{sec:synthetic}

In this first section, we provide a series of simulations on the estimation of $C$ and $C^{-1}$ based on the Fisher distance and for several examples of genuine matrices $C$. Similar results and conclusions were obtained for the other metrics discussed above (the KL divergence and the square Battacharrya distance especially) which are thus not explicitly reported here. The interested reader can refer to the code provided by the authors for self experimentation as supplementary material.

Our preference for the Fisher distance for fair comparisons lies in the fact that it is  the ``natural'' Riemannian distance to compare covariance matrices in $S_n^{++}$, therefore in entire agreement with the proposed estimation strategy through gradient descents in $S_n^{++}$. Besides, for this specific case, Theorem~4 in \cite{smith2005covariance} establishes an exact and very straightforward formula for the Cramer-Rao bound (CRB) on \emph{unbiased} estimators of $C$. Although the compared estimators of $C$ are likely all biased and that $M_0$ initializations may by chance bring additional information disrupting a formally fair CRB comparison, the CRB at least provides an indicator of relevance of the estimators. 

\medskip

The examples of covariance matrix $C$ under consideration in the following are: \\
\noindent(i) [Wishart] a random ($p$-dimensional) standard Wishart matrix with $2p$ degrees of freedom,\\
\noindent(ii) [Toeplitz $a$] the Toeplitz matrix defined by $C_{ij}=a^{|i-j|}$,\\
\noindent(iii) [Discrete] a matrix $C$ with uniform eigenvector distribution and eigenvalues equal to $.1$, $1$, $3$, $4$ each with multiplicity $p/4$.

\medskip

\begin{figure}[t]
\centering
\begin{tabular}{cc}
\hspace{-0.25cm}
    \begin{tikzpicture}[scale=.51]
    		\begin{semilogyaxis}[]
    			            \addplot[thin,mark=diamond,mark size=3pt,blue,thick]coordinates{                    (1.052632e+00,3.621215e+00)(1.226316e+00,2.155288e+00)(1.400000e+00,1.570634e+00)(1.578947e+00,1.231576e+00)(1.752632e+00,1.018700e+00)(1.926316e+00,8.684725e-01)(2.105263e+00,7.537480e-01)(2.278947e+00,6.679420e-01)(2.452632e+00,5.995629e-01)(2.631579e+00,5.422702e-01)

};
    			            \addplot[thin,mark=o,mark size=3pt,green,thick]coordinates{                    (1.052632e+00,3.694213e+00)(1.226316e+00,2.186403e+00)(1.400000e+00,1.578760e+00)(1.578947e+00,1.243065e+00)(1.752632e+00,1.028860e+00)(1.926316e+00,8.768624e-01)(2.105263e+00,7.595774e-01)(2.278947e+00,6.714819e-01)(2.452632e+00,6.038123e-01)(2.631579e+00,5.470798e-01)

};
    			            \addplot[thin,mark=square,mark size=3pt,red,thick]coordinates{                (1.052632e+00,1.630722e+00)(1.226316e+00,1.321043e+00)(1.400000e+00,1.132410e+00)(1.578947e+00,9.934601e-01)(1.752632e+00,8.922964e-01)(1.926316e+00,8.092820e-01)(2.105263e+00,7.423678e-01)(2.278947e+00,6.863085e-01)(2.452632e+00,6.384575e-01)(2.631579e+00,5.968495e-01)

};
    			            \addplot[thin,mark=star,mark size=3pt,orange,thick]coordinates{                (1.052632e+00,9.550000e-01)(1.226316e+00,8.197425e-01)(1.400000e+00,7.180451e-01)(1.578947e+00,6.366667e-01)(1.752632e+00,5.735736e-01)(1.926316e+00,5.218579e-01)(2.105263e+00,4.775000e-01)(2.278947e+00,4.411085e-01)(2.452632e+00,4.098712e-01)(2.631579e+00,3.820000e-01)

};
    			  \addplot[thin,mark=triangle,mark size=3pt,black,thick]coordinates{                (1.052632e+00,1.499977e+00)(1.226316e+00,1.287635e+00)(1.400000e+00,1.177093e+00)(1.578947e+00,1.047568e+00)(1.752632e+00,9.692024e-01)(1.926316e+00,8.744637e-01)(2.105263e+00,8.238015e-01)(2.278947e+00,7.436929e-01)(2.452632e+00,7.201485e-01)(2.631579e+00,6.477400e-01)

};
    			  \addplot[thin,mark=+,mark size=3pt,cyan,thick]coordinates{                    (1.052632e+00,1.271110e+00)(1.226316e+00,1.095364e+00)(1.400000e+00,1.011359e+00)(1.578947e+00,9.061224e-01)(1.752632e+00,8.429917e-01)(1.926316e+00,7.604303e-01)(2.105263e+00,7.209962e-01)(2.278947e+00,6.513282e-01)(2.452632e+00,6.333660e-01)(2.631579e+00,5.685213e-01)

};
    		\end{semilogyaxis}
    \end{tikzpicture}& 
    \hspace{-0.5cm}
    \begin{tikzpicture}[scale=.51]
            \begin{semilogyaxis}[]
    			            \addplot[ultra thin,mark=diamond,mark size=3pt,blue,thick]coordinates{                                        (1.052632e+00,3.621215e+00)(1.226316e+00,2.155288e+00)(1.400000e+00,1.570634e+00)(1.578947e+00,1.231576e+00)(1.752632e+00,1.018700e+00)(1.926316e+00,8.684725e-01)(2.105263e+00,7.537480e-01)(2.278947e+00,6.679420e-01)(2.452632e+00,5.995629e-01)(2.631579e+00,5.422702e-01)

};
    			            \addplot[ultra thin,mark=o,mark size=3pt,green,thick]coordinates{                                        (1.052632e+00,3.674975e+00)(1.226316e+00,2.176235e+00)(1.400000e+00,1.583820e+00)(1.578947e+00,1.242327e+00)(1.752632e+00,1.027019e+00)(1.926316e+00,8.773549e-01)(2.105263e+00,7.589141e-01)(2.278947e+00,6.728323e-01)(2.452632e+00,6.039138e-01)(2.631579e+00,5.472911e-01)

};
    			            \addplot[ultra thin,mark=square,mark size=3pt,red,thick]coordinates{                                        (1.052632e+00,1.188006e-01)(1.226316e+00,8.885209e-02)(1.400000e+00,7.222225e-02)(1.578947e+00,6.080059e-02)(1.752632e+00,5.278553e-02)(1.926316e+00,4.673874e-02)(2.105263e+00,4.241557e-02)(2.278947e+00,3.875348e-02)(2.452632e+00,3.610528e-02)(2.631579e+00,3.369227e-02)

};
    			            \addplot[ultra thin,mark=star,mark size=3pt,orange,thick]coordinates{                                        (1.052632e+00,9.550000e-01)(1.226316e+00,8.197425e-01)(1.400000e+00,7.180451e-01)(1.578947e+00,6.366667e-01)(1.752632e+00,5.735736e-01)(1.926316e+00,5.218579e-01)(2.105263e+00,4.775000e-01)(2.278947e+00,4.411085e-01)(2.452632e+00,4.098712e-01)(2.631579e+00,3.820000e-01)

};
\addplot[ultra thin,mark=triangle,mark size=3pt,black,thick]coordinates{                                   (1.052632e+00,1.994428e-02)(1.226316e+00,1.985250e-02)(1.400000e+00,1.994762e-02)(1.578947e+00,1.972225e-02)(1.752632e+00,1.969931e-02)(1.926316e+00,1.961005e-02)(2.105263e+00,1.946746e-02)(2.278947e+00,1.941751e-02)(2.452632e+00,1.929310e-02)(2.631579e+00,1.927338e-02)

};
\addplot[ultra thin,mark=+,mark size=3pt,cyan,thick]coordinates{                                   (1.052632e+00,2.067627e-02)(1.226316e+00,2.052808e-02)(1.400000e+00,2.137548e-02)(1.578947e+00,2.024707e-02)(1.752632e+00,2.045096e-02)(1.926316e+00,1.998795e-02)(2.105263e+00,1.975263e-02)(2.278947e+00,1.978745e-02)(2.452632e+00,1.952595e-02)(2.631579e+00,1.956909e-02)

};
    		\end{semilogyaxis}
    \end{tikzpicture}\\
     \vspace{-0.3cm}
    \hspace{-0.25cm}
    \begin{tikzpicture}[scale=.51]
            \begin{semilogyaxis}[xlabel=$\frac np$]
    			            \addplot[ultra thin,mark=diamond,mark size=3pt,blue,thick]coordinates{                        (1.052632e+00,3.621215e+00)(1.226316e+00,2.155288e+00)(1.400000e+00,1.570634e+00)(1.578947e+00,1.231576e+00)(1.752632e+00,1.018700e+00)(1.926316e+00,8.684725e-01)(2.105263e+00,7.537480e-01)(2.278947e+00,6.679420e-01)(2.452632e+00,5.995629e-01)(2.631579e+00,5.422702e-01)

};
    			            \addplot[ultra thin,mark=o,mark size=3pt,green,thick]coordinates{                        (1.052632e+00,3.694358e+00)(1.226316e+00,2.178335e+00)(1.400000e+00,1.583750e+00)(1.578947e+00,1.241910e+00)(1.752632e+00,1.028330e+00)(1.926316e+00,8.747156e-01)(2.105263e+00,7.601444e-01)(2.278947e+00,6.737765e-01)(2.452632e+00,6.035557e-01)(2.631579e+00,5.460528e-01)

};
    			            \addplot[ultra thin,mark=square,mark size=3pt,red,thick]coordinates{                    (1.052632e+00,7.371905e-01)(1.226316e+00,6.727938e-01)(1.400000e+00,6.169036e-01)(1.578947e+00,5.684583e-01)(1.752632e+00,5.286577e-01)(1.926316e+00,4.948597e-01)(2.105263e+00,4.635910e-01)(2.278947e+00,4.378237e-01)(2.452632e+00,4.160172e-01)(2.631579e+00,3.954926e-01)

};
    			            \addplot[ultra thin,mark=star,mark size=3pt,orange,thick]coordinates{                    (1.052632e+00,9.550000e-01)(1.226316e+00,8.197425e-01)(1.400000e+00,7.180451e-01)(1.578947e+00,6.366667e-01)(1.752632e+00,5.735736e-01)(1.926316e+00,5.218579e-01)(2.105263e+00,4.775000e-01)(2.278947e+00,4.411085e-01)(2.452632e+00,4.098712e-01)(2.631579e+00,3.820000e-01)

};
\addplot[ultra thin,mark=triangle,mark size=3pt,black,thick]coordinates{                        (1.052632e+00,6.396125e-01)(1.226316e+00,5.631075e-01)(1.400000e+00,5.043909e-01)(1.578947e+00,4.554750e-01)(1.752632e+00,4.135436e-01)(1.926316e+00,3.822715e-01)(2.105263e+00,3.546627e-01)(2.278947e+00,3.294523e-01)(2.452632e+00,3.087347e-01)(2.631579e+00,2.905306e-01)

};
\addplot[ultra thin,mark=+,mark size=3pt,cyan,thick]coordinates{                            (1.052632e+00,5.546988e-01)(1.226316e+00,4.984043e-01)(1.400000e+00,4.515672e-01)(1.578947e+00,4.132313e-01)(1.752632e+00,3.789653e-01)(1.926316e+00,3.541447e-01)(2.105263e+00,3.300938e-01)(2.278947e+00,3.094084e-01)(2.452632e+00,2.911783e-01)(2.631579e+00,2.749747e-01)

};
    		\end{semilogyaxis}
    \end{tikzpicture}&
    \hspace{-0.5cm}
    \begin{tikzpicture}[scale=.51]
            \begin{semilogyaxis}[xlabel=$\frac np$]
    			            \addplot[ultra thin,mark=diamond,mark size=3pt,blue,thick]coordinates{                            (1.052632e+00,3.621215e+00)(1.226316e+00,2.155288e+00)(1.400000e+00,1.570634e+00)(1.578947e+00,1.231576e+00)(1.752632e+00,1.018700e+00)(1.926316e+00,8.684725e-01)(2.105263e+00,7.537480e-01)(2.278947e+00,6.679420e-01)(2.452632e+00,5.995629e-01)(2.631579e+00,5.422702e-01)

};
    			            \addplot[ultra thin,mark=o,mark size=3pt,green,thick]coordinates{                            (1.052632e+00,3.697454e+00)(1.226316e+00,2.179000e+00)(1.400000e+00,1.585799e+00)(1.578947e+00,1.242235e+00)(1.752632e+00,1.026134e+00)(1.926316e+00,8.768157e-01)(2.105263e+00,7.591115e-01)(2.278947e+00,6.725637e-01)(2.452632e+00,6.056003e-01)(2.631579e+00,5.455430e-01)

};
    			            \addplot[ultra thin,mark=square,mark size=3pt,red,thick]coordinates{                        (1.052632e+00,7.253566e-01)(1.226316e+00,6.327986e-01)(1.400000e+00,5.641260e-01)(1.578947e+00,5.085930e-01)(1.752632e+00,4.649982e-01)(1.926316e+00,4.287137e-01)(2.105263e+00,3.968448e-01)(2.278947e+00,3.702480e-01)(2.452632e+00,3.472137e-01)(2.631579e+00,3.259028e-01)

};
    			            \addplot[ultra thin,mark=star,mark size=3pt,orange,thick]coordinates{                        (1.052632e+00,9.550000e-01)(1.226316e+00,8.197425e-01)(1.400000e+00,7.180451e-01)(1.578947e+00,6.366667e-01)(1.752632e+00,5.735736e-01)(1.926316e+00,5.218579e-01)(2.105263e+00,4.775000e-01)(2.278947e+00,4.411085e-01)(2.452632e+00,4.098712e-01)(2.631579e+00,3.820000e-01)

};
\addplot[ultra thin,mark=triangle,mark size=3pt,black,thick]coordinates{                (1.052632e+00,1.060180e+00)(1.226316e+00,8.121410e-01)(1.400000e+00,6.499538e-01)(1.578947e+00,5.480851e-01)(1.752632e+00,4.784629e-01)(1.926316e+00,4.270973e-01)(2.105263e+00,3.845834e-01)(2.278947e+00,3.504938e-01)(2.452632e+00,3.217629e-01)(2.631579e+00,2.967711e-01)

};
\addplot[ultra thin,mark=+,mark size=3pt,cyan,thick]coordinates{                    (1.052632e+00,1.339840e+00)(1.226316e+00,8.530916e-01)(1.400000e+00,6.445148e-01)(1.578947e+00,5.434726e-01)(1.752632e+00,4.747840e-01)(1.926316e+00,4.246852e-01)(2.105263e+00,3.829447e-01)(2.278947e+00,3.488174e-01)(2.452632e+00,3.199291e-01)(2.631579e+00,2.953459e-01)

};
    			            \legend{SCM,SCM th,Proposed,CRB,QuEST1,QuEST2}
    		\end{semilogyaxis}
    \end{tikzpicture}
\end{tabular}
\caption{Fisher distance of estimates of $C$, initialized at linear-shrinkage. From top-left to bottom-right: Wishart, Toeplitz $0.1$, Toeplitz $0.9$, Discrete. {\rm ``SCM th''} defined in Remark~\ref{rem:delta_C_hatC}. Averaged over $100$ random realizations of $X$, $p=200$.}
\label{fig:C_Fisher}
\end{figure}
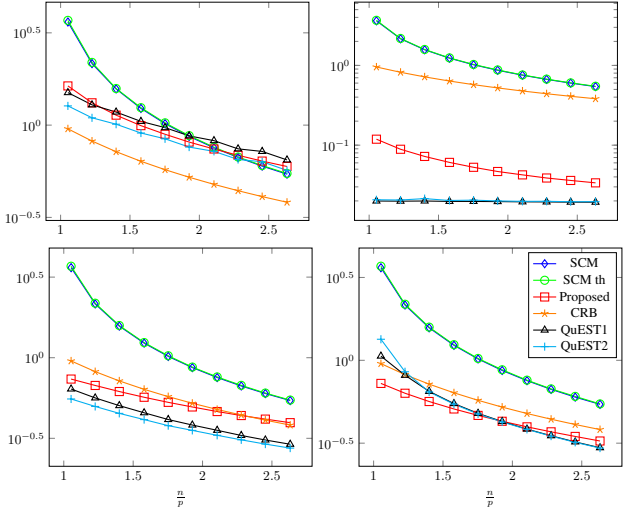

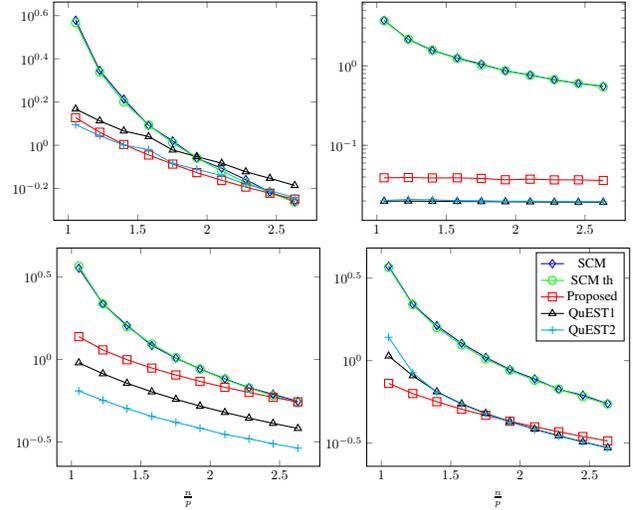
\begin{figure}[t]
\begin{tabular}{cc}
\hspace{-0.25cm}
    \begin{tikzpicture}[scale=0.51]
    		\begin{semilogyaxis}[]
    			            \addplot[thin,mark=diamond,mark size=3pt,blue,thick]coordinates{                                            (1.052632e+00,3.788769e+00)(1.226316e+00,2.217939e+00)(1.400000e+00,1.635436e+00)(1.578947e+00,1.235143e+00)(1.752632e+00,1.046948e+00)(1.926316e+00,8.699400e-01)(2.105263e+00,7.826091e-01)(2.278947e+00,6.932245e-01)(2.452632e+00,6.072428e-01)(2.631579e+00,5.478288e-01)

};
    			            \addplot[thin,mark=o,mark size=3pt,green,thick]coordinates{                                                (1.052632e+00,3.689722e+00)(1.226316e+00,2.181395e+00)(1.400000e+00,1.588211e+00)(1.578947e+00,1.240366e+00)(1.752632e+00,1.026280e+00)(1.926316e+00,8.739656e-01)(2.105263e+00,7.605838e-01)(2.278947e+00,6.732298e-01)(2.452632e+00,6.041772e-01)(2.631579e+00,5.453165e-01)

};
    			            \addplot[thin,mark=square,mark size=3pt,red,thick]coordinates{                                            (1.052632e+00,1.340825e+00)(1.226316e+00,1.148784e+00)(1.400000e+00,1.007940e+00)(1.578947e+00,9.008964e-01)(1.752632e+00,8.178906e-01)(1.926316e+00,7.470751e-01)(2.105263e+00,6.883630e-01)(2.278947e+00,6.405346e-01)(2.452632e+00,5.998823e-01)(2.631579e+00,5.631572e-01)

};
    			            \addplot[thin,mark=triangle,mark size=3pt,black,thick]coordinates{                                                (1.052632e+00,1.471466e+00)(1.226316e+00,1.296856e+00)(1.400000e+00,1.164339e+00)(1.578947e+00,1.097235e+00)(1.752632e+00,9.503871e-01)(1.926316e+00,8.852141e-01)(2.105263e+00,8.260845e-01)(2.278947e+00,7.538635e-01)(2.452632e+00,7.030142e-01)(2.631579e+00,6.512554e-01)

};
    			            \addplot[thin,mark=+,mark size=3pt,cyan,thick]coordinates{                                                    (1.052632e+00,1.245585e+00)(1.226316e+00,1.106857e+00)(1.400000e+00,1.001431e+00)(1.578947e+00,9.542531e-01)(1.752632e+00,8.244512e-01)(1.926316e+00,7.725414e-01)(2.105263e+00,7.244969e-01)(2.278947e+00,6.597331e-01)(2.452632e+00,6.167584e-01)(2.631579e+00,5.721614e-01)

};
    		\end{semilogyaxis}
    \end{tikzpicture}& 
    \hspace{-0.5cm}
    \begin{tikzpicture}[scale=0.51]
            \begin{semilogyaxis}[]
    			            \addplot[ultra thin,mark=diamond,mark size=3pt,blue,thick]coordinates{                            (1.052632e+00,3.741016e+00)(1.226316e+00,2.156684e+00)(1.400000e+00,1.563750e+00)(1.578947e+00,1.259255e+00)(1.752632e+00,1.054680e+00)(1.926316e+00,8.648557e-01)(2.105263e+00,7.676727e-01)(2.278947e+00,6.691129e-01)(2.452632e+00,6.008814e-01)(2.631579e+00,5.540627e-01)

};
    			            \addplot[ultra thin,mark=o,mark size=3pt,green,thick]coordinates{                                  (1.052632e+00,3.685598e+00)(1.226316e+00,2.172202e+00)(1.400000e+00,1.584984e+00)(1.578947e+00,1.241869e+00)(1.752632e+00,1.029069e+00)(1.926316e+00,8.720556e-01)(2.105263e+00,7.595110e-01)(2.278947e+00,6.724536e-01)(2.452632e+00,6.024094e-01)(2.631579e+00,5.452767e-01)

};
    			            \addplot[ultra thin,mark=square,mark size=3pt,red,thick]coordinates{                             (1.052632e+00,3.908628e-02)(1.226316e+00,3.940295e-02)(1.400000e+00,3.891659e-02)(1.578947e+00,3.905542e-02)(1.752632e+00,3.834711e-02)(1.926316e+00,3.691481e-02)(2.105263e+00,3.758303e-02)(2.278947e+00,3.681378e-02)(2.452632e+00,3.685479e-02)(2.631579e+00,3.619077e-02)

};
    			            \addplot[ultra thin,mark=triangle,mark size=3pt,black,thick]coordinates{                               (1.052632e+00,1.989210e-02)(1.226316e+00,1.993774e-02)(1.400000e+00,1.981406e-02)(1.578947e+00,1.971776e-02)(1.752632e+00,1.963929e-02)(1.926316e+00,1.955115e-02)(2.105263e+00,1.947210e-02)(2.278947e+00,1.936785e-02)(2.452632e+00,1.932336e-02)(2.631579e+00,1.926517e-02)

};
    			            \addplot[ultra thin,mark=+,mark size=3pt,cyan,thick]coordinates{                                   (1.052632e+00,2.030259e-02)(1.226316e+00,2.091271e-02)(1.400000e+00,2.072259e-02)(1.578947e+00,2.013384e-02)(1.752632e+00,2.011159e-02)(1.926316e+00,1.984391e-02)(2.105263e+00,1.986455e-02)(2.278947e+00,1.965488e-02)(2.452632e+00,1.953615e-02)(2.631579e+00,1.955380e-02)

};
    		\end{semilogyaxis}
    \end{tikzpicture}\\
     \vspace{-0.3cm}
    \hspace{-0.25cm}
    \begin{tikzpicture}[scale=0.51]
            \begin{semilogyaxis}[xlabel=$\frac np$]
    			            \addplot[ultra thin,mark=diamond,mark size=3pt,blue,thick]coordinates{                                    (1.052632e+00,3.579181e+00)(1.226316e+00,2.175304e+00)(1.400000e+00,1.611753e+00)(1.578947e+00,1.215484e+00)(1.752632e+00,1.019890e+00)(1.926316e+00,8.764071e-01)(2.105263e+00,7.646716e-01)(2.278947e+00,6.744528e-01)(2.452632e+00,6.150928e-01)(2.631579e+00,5.572720e-01)

};
    			            \addplot[ultra thin,mark=o,mark size=3pt,green,thick]coordinates{                                    (1.052632e+00,3.680204e+00)(1.226316e+00,2.175567e+00)(1.400000e+00,1.584603e+00)(1.578947e+00,1.241733e+00)(1.752632e+00,1.025340e+00)(1.926316e+00,8.755925e-01)(2.105263e+00,7.589804e-01)(2.278947e+00,6.724884e-01)(2.452632e+00,6.040264e-01)(2.631579e+00,5.464114e-01)

};
    			            \addplot[ultra thin,mark=square,mark size=3pt,red,thick]coordinates{                                (1.052632e+00,1.375504e+00)(1.226316e+00,1.143196e+00)(1.400000e+00,9.969414e-01)(1.578947e+00,8.877479e-01)(1.752632e+00,8.056267e-01)(1.926316e+00,7.363836e-01)(2.105263e+00,6.782199e-01)(2.278947e+00,6.313230e-01)(2.452632e+00,5.898757e-01)(2.631579e+00,5.534163e-01)

};
    			            \addplot[ultra thin,mark=triangle,mark size=3pt,black,thick]coordinates{                                    (1.052632e+00,9.550000e-01)(1.226316e+00,8.197425e-01)(1.400000e+00,7.180451e-01)(1.578947e+00,6.366667e-01)(1.752632e+00,5.735736e-01)(1.926316e+00,5.218579e-01)(2.105263e+00,4.775000e-01)(2.278947e+00,4.411085e-01)(2.452632e+00,4.098712e-01)(2.631579e+00,3.820000e-01)

};
    			            \addplot[ultra thin,mark=+,mark size=3pt,cyan,thick]coordinates{                                        (1.052632e+00,6.441574e-01)(1.226316e+00,5.656338e-01)(1.400000e+00,5.039669e-01)(1.578947e+00,4.517752e-01)(1.752632e+00,4.154066e-01)(1.926316e+00,3.831613e-01)(2.105263e+00,3.512833e-01)(2.278947e+00,3.302324e-01)(2.452632e+00,3.083451e-01)(2.631579e+00,2.900795e-01)

};
    		\end{semilogyaxis}
    \end{tikzpicture}&
    \hspace{-0.5cm}
    \begin{tikzpicture}[scale=0.51]
            \begin{semilogyaxis}[xlabel=$\frac np$]
    			            \addplot[thin,mark=diamond,mark size=3pt,blue,thick]coordinates{                                (1.052632e+00,3.735637e+00)(1.226316e+00,2.199609e+00)(1.400000e+00,1.628725e+00)(1.578947e+00,1.274215e+00)(1.752632e+00,1.046850e+00)(1.926316e+00,8.839321e-01)(2.105263e+00,7.748881e-01)(2.278947e+00,6.687536e-01)(2.452632e+00,6.174820e-01)(2.631579e+00,5.484381e-01)

};
    			            \addplot[thin,mark=o,mark size=3pt,green,thick]coordinates{                                (1.052632e+00,3.679596e+00)(1.226316e+00,2.181651e+00)(1.400000e+00,1.586045e+00)(1.578947e+00,1.244708e+00)(1.752632e+00,1.024387e+00)(1.926316e+00,8.751126e-01)(2.105263e+00,7.608659e-01)(2.278947e+00,6.730758e-01)(2.452632e+00,6.049433e-01)(2.631579e+00,5.457421e-01)

};
    			            \addplot[thin,mark=square,mark size=3pt,red,thick]coordinates{                                (1.052632e+00,7.281018e-01)(1.226316e+00,6.315875e-01)(1.400000e+00,5.640325e-01)(1.578947e+00,5.071892e-01)(1.752632e+00,4.662584e-01)(1.926316e+00,4.289031e-01)(2.105263e+00,3.971919e-01)(2.278947e+00,3.706304e-01)(2.452632e+00,3.467929e-01)(2.631579e+00,3.261004e-01)

};
    			            \addplot[thin,mark=triangle,mark size=3pt,black,thick]coordinates{                                (1.052632e+00,1.067290e+00)(1.226316e+00,8.102669e-01)(1.400000e+00,6.481152e-01)(1.578947e+00,5.470527e-01)(1.752632e+00,4.792370e-01)(1.926316e+00,4.269285e-01)(2.105263e+00,3.844490e-01)(2.278947e+00,3.506594e-01)(2.452632e+00,3.217566e-01)(2.631579e+00,2.967694e-01)

};
    			            \addplot[thin,mark=+,mark size=3pt,cyan,thick]coordinates{                                    (1.052632e+00,1.386300e+00)(1.226316e+00,8.444241e-01)(1.400000e+00,6.432153e-01)(1.578947e+00,5.434294e-01)(1.752632e+00,4.753575e-01)(1.926316e+00,4.250168e-01)(2.105263e+00,3.825568e-01)(2.278947e+00,3.484675e-01)(2.452632e+00,3.208061e-01)(2.631579e+00,2.951085e-01)

};
    			            \legend{SCM,SCM th,Proposed,QuEST1,QuEST2}
    		\end{semilogyaxis}
    \end{tikzpicture}
\end{tabular}
\caption{Fisher distance of estimates of $C^{-1}$, initialized at linear-shrinkage. From top-left to bottom-right: Wishart, Toeplitz $0.1$, Toeplitz $0.9$, Discrete. Averaged over $100$ random realizations of $X$, $p=200$.}
\label{fig:Cinv_Fisher}
\end{figure}

Figure~\ref{fig:C_Fisher} (for the estimation of $C$) and Figure~\ref{fig:Cinv_Fisher} (for $C^{-1}$) report comparative performances on the aforementioned $C$ matrices for the SCM, QuEST1, QuEST2, and our proposed estimator, the latter three being initialized at $M_0$ the shrinkage estimation from \cite{LED04} (consistent with the choice made for QuEST1, QuEST2 in \cite{LW15,LW18}). In the figures, ``{\rm SCM th}'' refers to the asymptotic analytical approximation of $\delta(C,\hat{C})$ as defined in Remark~\ref{rem:delta_C_hatC}. 
It is observed that, while the SCM never reaches the unbiased CRB, in many cases the QuESTx estimators and our proposed method overtake the CRB, sometimes significantly so. The Wishart matrix case seems more challenging from this perspective. 
In terms of performances, both our proposed method and QuESTx perform competitively and systematically better than the sample covariance matrix.

\begin{remark}[Consistent estimator for $\delta(C,\hat{C})$]
\label{rem:delta_C_hatC}
With the same technical tools from \cite{couillet2018random}, it is straightforward to estimate the distance $\delta(C,\hat{C})$. Indeed, $\delta(C,\hat{C})=\frac1{2\pi\imath}\oint_{\Gamma} f(z)m_{\gamma}(z)$ for $m_{\gamma}(z)$ the Stieljes transform of the eigenvalue distribution of $C^{-1}\hat{C}$; the limiting distribution of the latter is the popular Marcenko-Pastur law \cite{MAR67}, the expression of which is well known. The estimate is denoted ``{\rm SCM th}'' in Figures~\ref{fig:C_Fisher}--\ref{fig:Cinv_Fisher}. The observed perfect match between limiting theory and practice confirms the consistency of the random matrix approach even for not too large $p,n$.
\end{remark}

\vspace{-.3cm}

\subsection{Application to LDA/QDA}
As pointed out in the introduction, the estimation of the covariance and inverse covariance matrices of random vectors are at the core of a wide range of applications in statistics, machine learning and signal processing. As a basic illustrative example, we focus here on linear discriminant analysis (LDA) and quadratic discriminant analysis (QDA). Both exploit estimates covariance matrices of the data or their inverse in order to perform the classification. 

Suppose $x^{(1)}_1,\ldots,x^{(1)}_{n_1}\sim N(\mu_1,C_1)$ and $x^{(2)}_1,\ldots,x^{(2)}_{n_2}\sim N(\mu_2,C_2)$ are two sets of random independent $p$-dimensional training vectors forming two classes of a Gaussian mixture. The objective of LDA and QDA is to estimate the probability for an arbitrary random vector $x$ to belong to either class by replacing the genuine means $\mu_a$ and covariances $C_a$ by sample estimates, with in the case of LDA the underlying (possibly erroneous) assumption that $C_1=C_2$. Defining $C$ as $C\equiv\frac{n_1}{n_1+n_2}C_1+\frac{n_2}{n_1+n_2}C_2$, the classification rules for LDA and QDA for data point $x$ depend on the signs of the respective quantities:
\begin{align*}
    &\delta_{x}^{\rm LDA}=(\hat \mu_1-\hat \mu_2)^\trans \check C^{-1}x+\frac12\mu_2^\trans \check C^{-1}\mu_2-\frac12\hat \mu_1^\trans \check C^{-1}\hat \mu_1\\
    &\delta_{x}^{\rm QDA}=\frac 12 x^\trans\left(\check C_2^{-1}-\check C_1^{-1}\right)x+\left(\hat\mu_1^\trans \check C_1^{-1}-\hat \mu_2^\trans \check C_2^{-1}\right)x\\
    &+\frac{1}{2}\hat\mu_2^\trans \check C_{2}^{-1}\hat\mu_2-\frac{1}{2}\hat \mu_1^\trans \check C_{1}^{-1}\hat\mu_1+\frac12\log {\rm det}\frac{\check C_1^{-1}}{\check C_2^{-1}}-\log\frac{n_2}{n_1}
\end{align*}
where $\hat\mu_a\equiv\frac1{n_a}\sum_{i=1}^{n_a}x_i^{(a)}$ is the sample estimate of $\mu_a$ and $\check C_a^{-1}$ are some estimate of $C_a^{-1}$, while $\check C\equiv \frac{n_1}{n_1+n_2}\check C_1+\frac{n_2}{n_1+n_2}\check C_2$ with $\check C_a$ the estimation of $C_a$. As such, in the following simulations, LDA will exclusively exploit estimations of $C_1$ and $C_2$ (before inverting their estimated average), while QDA will focus on estimating directly the inverses $C_1^{-1}$ and $C_2^{-1}$.

The first three displays in Figures~\ref{fig:LDA_performance} and \ref{fig:QDA_performance} compare the accuracies of the LDA/QDA algorithms for $C_1$ and $C_2$ chosen among Wishart and Toeplitz matrices, and for $\mu_2=\mu_1+\frac{80}p$ for the LDA and $\mu_2=\mu_1+\frac1p$ for the QDA settings (in order to avoid trivial classification). The bottom right displays are applications to real EEG data extracted from the dataset in \cite{andrzejak2001indications}. The dataset contains five subsets (denoted A-E). Sets A and B were collected from healthy volunteers while C, D, E were collected from epileptic  patients. The graph presents all combinations of binary classes between healthy volunteers and epilectic subjects (e.g., A/E for subsets A and E). There we observe that, for most considered settings, our proposed algorithm almost systematically outperforms competing methods, with QuEST1 and QuEST2 exhibiting a much less stable behavior and particularly weak performances in all synthetic scenarios.

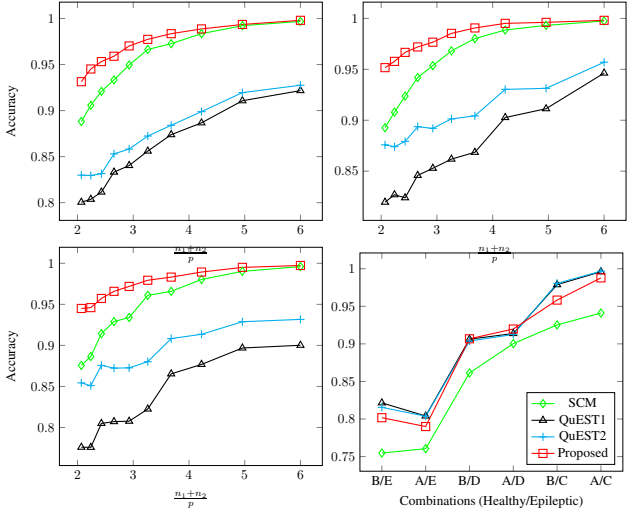
\begin{figure}[t]
\begin{tabular}{cc}
\hspace{-0.25cm}
    \begin{tikzpicture}[scale=0.51]
    		\begin{axis}[legend pos= south east,ylabel=Accuracy,xlabel=$\frac {n_1+n_2}p$]
    			            \addplot[thin,mark=diamond,mark size=3pt,green,thick]coordinates{                       (2.068966e+00,8.881667e-01)(2.238806e+00,9.056667e-01)(2.429150e+00,9.208333e-01)(2.654867e+00,9.333333e-01)(2.926829e+00,9.495000e-01)(3.260870e+00,9.663333e-01)(3.680982e+00,9.726667e-01)(4.225352e+00,9.835000e-01)(4.958678e+00,9.921667e-01)(6,9.968333e-01)

};
    			            \addplot[thin,mark=triangle,mark size=3pt,black,thick]coordinates{                  
        (2.068966e+00,8.005000e-01)(2.238806e+00,8.036667e-01)(2.429150e+00,8.115000e-01)(2.654867e+00,8.331667e-01)(2.926829e+00,8.403333e-01)(3.260870e+00,8.560000e-01)(3.680982e+00,8.738333e-01)(4.225352e+00,8.866667e-01)(4.958678e+00,9.106667e-01)(6,9.216667e-01)

};
    			            \addplot[thin,mark=+,mark size=3pt,cyan,thick]coordinates{                  (2.068966e+00,8.298333e-01)(2.238806e+00,8.295000e-01)(2.429150e+00,8.315000e-01)(2.654867e+00,8.530000e-01)(2.926829e+00,8.581667e-01)(3.260870e+00,8.723333e-01)(3.680982e+00,8.840000e-01)(4.225352e+00,8.988333e-01)(4.958678e+00,9.195000e-01)(6,9.275000e-01)
};
    			            \addplot[thin,mark=square,mark size=3pt,red,thick]coordinates{                      (2.068966e+00,9.313333e-01)(2.238806e+00,9.451667e-01)(2.429150e+00,9.531667e-01)(2.654867e+00,9.590000e-01)(2.926829e+00,9.701667e-01)(3.260870e+00,9.773333e-01)(3.680982e+00,9.835000e-01)(4.225352e+00,9.886667e-01)(4.958678e+00,9.935000e-01)(6,9.980000e-01)

};

    		\end{axis}
    \end{tikzpicture}& 
     \vspace{-0.4cm}
    \hspace{-0.5cm}
    \begin{tikzpicture}[scale=0.51]
            \begin{axis}[legend pos= south east,xlabel=$\frac {n_1+n_2}p$]
    			            \addplot[ultra thin,mark=diamond,mark size=3pt,green,thick]coordinates{                (2.068966e+00,8.926667e-01)(2.238806e+00,9.080000e-01)(2.429150e+00,9.236667e-01)(2.654867e+00,9.420000e-01)(2.926829e+00,9.536667e-01)(3.260870e+00,9.681667e-01)(3.680982e+00,9.801667e-01)(4.225352e+00,9.886667e-01)(4.958678e+00,9.931667e-01)(6,9.975000e-01)

};
    			            \addplot[ultra thin,mark=triangle,mark size=3pt,black,thick]coordinates{                
                       (2.068966e+00,8.195000e-01)(2.238806e+00,8.268333e-01)(2.429150e+00,8.238333e-01)(2.654867e+00,8.458333e-01)(2.926829e+00,8.528333e-01)(3.260870e+00,8.618333e-01)(3.680982e+00,8.685000e-01)(4.225352e+00,9.026667e-01)(4.958678e+00,9.113333e-01)(6,9.463333e-01)

};
    			            \addplot[ultra thin,mark=+,mark size=3pt,cyan,thick]coordinates{                    (2.068966e+00,8.758333e-01)(2.238806e+00,8.740000e-01)(2.429150e+00,8.791667e-01)(2.654867e+00,8.936667e-01)(2.926829e+00,8.920000e-01)(3.260870e+00,9.013333e-01)(3.680982e+00,9.043333e-01)(4.225352e+00,9.303333e-01)(4.958678e+00,9.313333e-01)(6,9.570000e-01)

};
    			            \addplot[ultra thin,mark=square,mark size=3pt,red,thick]coordinates{                        (2.068966e+00,9.516667e-01)(2.238806e+00,9.578333e-01)(2.429150e+00,9.666667e-01)(2.654867e+00,9.720000e-01)(2.926829e+00,9.766667e-01)(3.260870e+00,9.853333e-01)(3.680982e+00,9.906667e-01)(4.225352e+00,9.951667e-01)(4.958678e+00,9.961667e-01)(6,9.981667e-01)

};
    		\end{axis}
    \end{tikzpicture}\\
    \hspace{-0.25cm}
    \begin{tikzpicture}[scale=0.51]
            \begin{axis}[legend pos= south east,xlabel=$\frac {n_1+n_2}p$,ylabel=Accuracy]
    			            \addplot[ultra thin,mark=diamond,mark size=3pt,green,thick]coordinates{                (2.068966e+00,8.756667e-01)(2.238806e+00,8.865000e-01)(2.429150e+00,9.143333e-01)(2.654867e+00,9.290000e-01)(2.926829e+00,9.340000e-01)(3.260870e+00,9.610000e-01)(3.680982e+00,9.658333e-01)(4.225352e+00,9.803333e-01)(4.958678e+00,9.901667e-01)(6,9.960000e-01)

};
    			            \addplot[ultra thin,mark=triangle,mark size=3pt,black,thick]coordinates{                      (2.068966e+00,7.760000e-01)(2.238806e+00,7.758333e-01)(2.429150e+00,8.050000e-01)(2.654867e+00,8.071667e-01)(2.926829e+00,8.076667e-01)(3.260870e+00,8.225000e-01)(3.680982e+00,8.653333e-01)(4.225352e+00,8.768333e-01)(4.958678e+00,8.968333e-01)(6,9.001667e-01)

};
    			            \addplot[ultra thin,mark=+,mark size=3pt,cyan,thick]coordinates{                      (2.068966e+00,8.543333e-01)(2.238806e+00,8.510000e-01)(2.429150e+00,8.756667e-01)(2.654867e+00,8.723333e-01)(2.926829e+00,8.726667e-01)(3.260870e+00,8.801667e-01)(3.680982e+00,9.083333e-01)(4.225352e+00,9.135000e-01)(4.958678e+00,9.288333e-01)(6,9.316667e-01)

};
    			            \addplot[ultra thin,mark=square,mark size=3pt,red,thick]coordinates{                          (2.068966e+00,9.450000e-01)(2.238806e+00,9.460000e-01)(2.429150e+00,9.571667e-01)(2.654867e+00,9.658333e-01)(2.926829e+00,9.718333e-01)(3.260870e+00,9.793333e-01)(3.680982e+00,9.831667e-01)(4.225352e+00,9.893333e-01)(4.958678e+00,9.950000e-01)(6,9.973333e-01)

};
    		\end{axis}
    \end{tikzpicture}&
    \hspace{-0.5cm}
    \begin{tikzpicture}[scale=0.51]
            \begin{axis}[legend pos= south east,xlabel=Combinations (Healthy/Epileptic),xtick=data,xticklabels={B/E,A/E,B/D,A/D,B/C,A/C}]
    			            \addplot[thin,mark=diamond,mark size=3pt,green,thick]coordinates{                      (1,7.547500e-01)(2,7.605000e-01)(3,8.612500e-01)(4,9.002500e-01)(5,9.252500e-01)(6,9.410000e-01)

};
    			            \addplot[thin,mark=triangle,mark size=3pt,black,thick]coordinates{                              (1,8.215000e-01)(2,8.040000e-01)(3,9.062500e-01)(4,9.137500e-01)(5,9.787500e-01)(6,9.960000e-01)

};
    			            \addplot[thin,mark=+,mark size=3pt,cyan,thick]coordinates{                             (1,8.155000e-01)(2,8.037500e-01)(3,9.037500e-01)(4,9.125000e-01)(5,9.805000e-01)(6,9.967500e-01)

};
    			            \addplot[thin,mark=square,mark size=3pt,red,thick]coordinates{                       (1,8.017500e-01)(2,7.900000e-01)(3,9.067500e-01)(4,9.197500e-01)(5,9.582500e-01)(6,9.877500e-01)

};
    			            \legend{SCM,QuEST1,QuEST2,Proposed}
    		\end{axis}
    \end{tikzpicture}
\end{tabular}
\caption{Mean accuracy obtained over $10$ realizations of LDA classification. From left to right and top to bottom: $C_1$ and $C_2$ are respectively Wishart/Wishart (independent), Wishart/Toeplitz-$0.2$, Toeplitz-$0.2$/Toeplitz-$0.4$, and real application to EEG data.}
\label{fig:LDA_performance}
\end{figure}
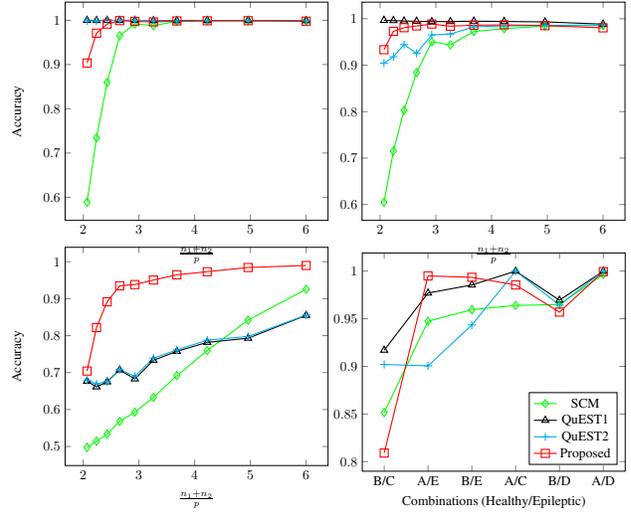
\begin{figure}[h]
\begin{tabular}{cc}
\hspace{-0.25cm}
    \begin{tikzpicture}[scale=0.51]
    		\begin{axis}[legend pos=south east,ylabel=Accuracy,xlabel=$\frac {n_1+n_2}p$]
    			            \addplot[thin,mark=diamond,mark size=3pt,green,thick]coordinates{                           (2.068966e+00,5.888333e-01)(2.238806e+00,7.343333e-01)(2.429150e+00,8.595000e-01)(2.654867e+00,9.651667e-01)(2.926829e+00,9.918333e-01)(3.260870e+00,9.880000e-01)(3.680982e+00,9.978333e-01)(4.225352e+00,9.993333e-01)(4.958678e+00,9.988333e-01)(6,9.991667e-01)

};
    			            \addplot[thin,mark=triangle,mark size=3pt,black,thick]coordinates{                              (2.068966e+00,1)(2.238806e+00,9.996667e-01)(2.429150e+00,1)(2.654867e+00,9.998333e-01)(2.926829e+00,1)(3.260870e+00,9.993333e-01)(3.680982e+00,9.991667e-01)(4.225352e+00,9.995000e-01)(4.958678e+00,9.996667e-01)(6,9.991667e-01)

};
    			            \addplot[thin,mark=+,mark size=3pt,cyan,thick]coordinates{                         
                (2.068966e+00,9.988333e-01)(2.238806e+00,9.990000e-01)(2.429150e+00,1)(2.654867e+00,9.998333e-01)(2.926829e+00,1)(3.260870e+00,9.993333e-01)(3.680982e+00,9.988333e-01)(4.225352e+00,9.995000e-01)(4.958678e+00,9.996667e-01)(6,9.991667e-01)

};
    			            \addplot[thin,mark=square,mark size=3pt,red,thick]coordinates{                         
                    (2.068966e+00,9.035000e-01)(2.238806e+00,9.710000e-01)(2.429150e+00,9.915000e-01)(2.654867e+00,9.993333e-01)(2.926829e+00,9.975000e-01)(3.260870e+00,9.963333e-01)(3.680982e+00,9.986667e-01)(4.225352e+00,9.990000e-01)(4.958678e+00,9.991667e-01)(6,9.980000e-01)

};

    		\end{axis}
    \end{tikzpicture}& 
    \vspace{-0.4cm}
    \hspace{-0.5cm}
    \begin{tikzpicture}[scale=0.51]
            \begin{axis}[legend pos=south east,xlabel=$\frac {n_1+n_2}p$]
    			            \addplot[ultra thin,mark=diamond,mark size=3pt,green,thick]coordinates{                    (2.068966e+00,6.050000e-01)(2.238806e+00,7.150000e-01)(2.429150e+00,8.031667e-01)(2.654867e+00,8.841667e-01)(2.926829e+00,9.503333e-01)(3.260870e+00,9.436667e-01)(3.680982e+00,9.725000e-01)(4.225352e+00,9.785000e-01)(4.958678e+00,9.841667e-01)(6,9.853333e-01)

};
			            \addplot[ultra thin,mark=triangle,mark size=3pt,black,thick]coordinates{                      (2.068966e+00,9.965000e-01)(2.238806e+00,9.960000e-01)(2.429150e+00,9.950000e-01)(2.654867e+00,9.945000e-01)(2.926829e+00,9.936667e-01)(3.260870e+00,9.938333e-01)(3.680982e+00,9.946667e-01)(4.225352e+00,9.940000e-01)(4.958678e+00,9.931667e-01)(6,9.883333e-01)

};
    			            \addplot[ultra thin,mark=+,mark size=3pt,cyan,thick]coordinates{                    
                        (2.068966e+00,9.043333e-01)(2.238806e+00,9.180000e-01)(2.429150e+00,9.443333e-01)(2.654867e+00,9.255000e-01)(2.926829e+00,9.651667e-01)(3.260870e+00,9.670000e-01)(3.680982e+00,9.826667e-01)(4.225352e+00,9.843333e-01)(4.958678e+00,9.855000e-01)(6,9.866667e-01)

};
    			            \addplot[ultra thin,mark=square,mark size=3pt,red,thick]coordinates{                    
                            (2.068966e+00,9.333333e-01)(2.238806e+00,9.728333e-01)(2.429150e+00,9.806667e-01)(2.654867e+00,9.845000e-01)(2.926829e+00,9.885000e-01)(3.260870e+00,9.838333e-01)(3.680982e+00,9.860000e-01)(4.225352e+00,9.866667e-01)(4.958678e+00,9.850000e-01)(6,9.803333e-01)

};
    		\end{axis}
    \end{tikzpicture}\\
    \hspace{-0.25cm}
    \begin{tikzpicture}[scale=0.51]
            \begin{axis}[legend pos=south east,xlabel=$\frac {n_1+n_2}p$,ylabel=Accuracy]
    			            \addplot[ultra thin,mark=diamond,mark size=3pt,green,thick]coordinates{                      (2.068966e+00,4.976667e-01)(2.238806e+00,5.143333e-01)(2.429150e+00,5.336667e-01)(2.654867e+00,5.680000e-01)(2.926829e+00,5.930000e-01)(3.260870e+00,6.330000e-01)(3.680982e+00,6.918333e-01)(4.225352e+00,7.595000e-01)(4.958678e+00,8.421667e-01)(6,9.253333e-01)

};
    			            \addplot[ultra thin,mark=triangle,mark size=3pt,black,thick]coordinates{                         (2.068966e+00,6.766667e-01)(2.238806e+00,6.613333e-01)(2.429150e+00,6.750000e-01)(2.654867e+00,7.071667e-01)(2.926829e+00,6.830000e-01)(3.260870e+00,7.333333e-01)(3.680982e+00,7.575000e-01)(4.225352e+00,7.818333e-01)(4.958678e+00,7.930000e-01)(6,8.550000e-01)

};
    			            \addplot[ultra thin,mark=+,mark size=3pt,cyan,thick]coordinates{                              (2.068966e+00,6.793333e-01)(2.238806e+00,6.700000e-01)(2.429150e+00,6.753333e-01)(2.654867e+00,7.081667e-01)(2.926829e+00,6.901667e-01)(3.260870e+00,7.391667e-01)(3.680982e+00,7.613333e-01)(4.225352e+00,7.878333e-01)(4.958678e+00,7.976667e-01)(6,8.556667e-01)

};
    			            \addplot[ultra thin,mark=square,mark size=3pt,red,thick]coordinates{                                  (2.068966e+00,7.041667e-01)(2.238806e+00,8.218333e-01)(2.429150e+00,8.915000e-01)(2.654867e+00,9.348333e-01)(2.926829e+00,9.381667e-01)(3.260870e+00,9.505000e-01)(3.680982e+00,9.646667e-01)(4.225352e+00,9.730000e-01)(4.958678e+00,9.843333e-01)(6,9.898333e-01)

};

    		\end{axis}
    \end{tikzpicture}&
    \hspace{-0.5cm}
    \begin{tikzpicture}[scale=0.51]
            \begin{axis}[legend pos=south east,xlabel=Combinations (Healthy/Epileptic),xtick=data,xticklabels={B/C,A/E,B/E,A/C,B/D,A/D}]
    			            \addplot[thin,mark=diamond,mark size=3pt,green,thick]coordinates{                      (1,8.515000e-01)(2,9.475000e-01)(3,9.595000e-01)(4,9.640000e-01)(5,9.650000e-01)(6,9.965000e-01)

};
    			            \addplot[thin,mark=triangle,mark size=3pt,black,thick]coordinates{                          (1,9.170000e-01)(2,9.770000e-01)(3,9.855000e-01)(4,1)(5,9.695000e-01)(6,1)

};
    			            \addplot[thin,mark=+,mark size=3pt,cyan,thick]coordinates{                      (1,9.020000e-01)(2,9.005000e-01)(3,9.435000e-01)(4,1)(5,9.645000e-01)(6,1)

};
    			            \addplot[thin,mark=square,mark size=3pt,red,thick]coordinates{                         (1,8.090000e-01)(2,9.950000e-01)(3,9.935000e-01)(4,9.855000e-01)(5,9.570000e-01)(6,9.995000e-01)

};
    			            \legend{SCM,QuEST1,QuEST2,Proposed}
    		\end{axis}
    \end{tikzpicture}
\end{tabular}
\caption{Mean accuracy obtained over $10$ realizations of QDA classification. From left to right and top to bottom: $C_1$ and $C_2$ are respectively Wishart/Wishart (independent), Wishart/Toeplitz-$0.2$, Toeplitz-$0.2$/Toeplitz-$0.4$, and real application to EEG data.}
\label{fig:QDA_performance}
\end{figure}

\section{Discussion and Concluding Remarks}
\label{sec:conclusion}
Based on elementary yet powerful contour integration techniques and random matrix theory, we have proposed in this work a systematic framework for the estimation of covariance and precision matrices. Unlike alternative state-of-the-art techniques that attempt to invert the fundamental Bai--Silverstein equations \cite{SIL95}, our proposed method relies on a basic gradient descent approach in $S_n^{++}$ that, in addition to performing competitively (if not better), is computationally simpler.

While restricted to metrics depending on the eigenvalues of products of covariance matrices, our approach may be flexibly adapted to further matrix divergences solely depending on eigenvalue relations. The same framework can notably be applied to the Wasserstein distance between zero-mean Gaussian laws. A reservation nonetheless remains on the need for ${n>p}$ in settings involving inverse matrices that, when not met, does not allow to relate the sought-for eigenvalues to the (undefined) inverse sample covariance matrices. \cite{couillet2018random} shows that this problem can be partially avoided for some divergences (not for the Fisher distance though). A systematic treatment of the ${n<p}$ case is however lacking.

\medskip

Our approach also suffers from profound theoretical limitations that need be properly addressed: the fact that $\hat{\delta}(M,X)$ only estimates the sought-for $\delta(M,C)$ for $M$ independent of $X$ poses a formal problem when implemented in the gradient descent approach. This needs be tackled: (i) either by estimating the introduced bias so to estimate the loss incurred or, better, (ii) by accounting for the dependence to provide a further estimator $\hat{\hat \delta}(M(X),X)$ of $\delta(M(X),C)$ for all $X$-\emph{dependent} matrices $M(X)$ following a specific form. Notably, given that, along the gradient descent initialized at $M_0=\hat UD_0\hat U^\trans$, with $\hat U$ the eigenvectors of $\hat C$ and $D_0$ some diagonal matrix, all subsequent $M_k$ matrices share the same profile (i.e., $M_k=\hat UD_k\hat U^\trans$ for some diagonal $D_k$), a first improvement would consist in estimating consistently $\delta(\hat UD\hat U^\trans,C)$ for deterministic diagonal matrices $D$.

\medskip

The stability of the eigenvectors when initialized at $M_0=\hat UD_0\hat U^\trans$, with $\hat U$ the eigenvectors of $\hat C$, turns our algorithm into a ``non-linear shrinkage'' method, as called by \cite{LW15}. Parallel simulations also suggest that arbitrary initializations $M_0$ which \emph{do not} follow this structure tend to still lead to solutions converging to the eigenspace of $\hat C$. However, this might be a consequence of the inconsistency of $\hat{\delta}(M,X)$ for $X$-dependent matrices $M$: it is expected that more consistent estimators might avoid this problem of \emph{``eigenvector of $\hat C$'' attraction}, thereby likely leading to improved estimations of the eigenvectors of $C$.

\medskip

We conclude by emphasizing that modern large dimensional statistics have lately realized that substituting large covariance matrices by their sample estimators (or even by improved covariance estimators) is in general a weak approach, and that one should rather focus on estimating some ultimate functional (e.g., the result of a statistical test) involving the covariance (see, e.g., \cite{MES08c} in array processing or \cite{YAN15} in statistical finance). It is to be noted that our proposed approach is consistent with these considerations as various functionals of $C$ can be obtained from Equation~\eqref{eq:Cauchy}, from which similar derivations can be performed.
\section*{Acknowledgement}
This work is supported by the ANR Project RMT4GRAPH (ANR-14-CE28-0006) and by the IDEX GSTATS Chair at University Grenoble Alpes.

\bibliographystyle{icml2019}
\bibliography{refs_Romain.bib}
\end{document}